\documentclass{article}

\usepackage{arxiv}

\usepackage[utf8]{inputenc} 
\usepackage[T1]{fontenc}    
\usepackage{hyperref}       
\usepackage{url}            
\usepackage{booktabs}       
\usepackage{multirow}
\usepackage{amsmath}
\usepackage{amsfonts}       
\usepackage{amssymb}
\usepackage{nicefrac}       
\usepackage{microtype}      
\usepackage{graphicx}
\usepackage{xcolor}
\usepackage{xspace}
\usepackage[most]{tcolorbox}   
\usepackage{fvextra}          
\graphicspath{ {./images/} }

\usepackage{setspace}
\usepackage{natbib}

\newcommand{\name}{MedRealMM\xspace}

\renewcommand{\headeright}{}
\renewcommand{\undertitle}{}

\title{MedRealMM: A Real-World Multimodal Benchmark for Chinese Online Medical Consultation}
\date{}

\author{
  Runhan Shi$^{1,2,\dagger,*}$ \quad Quan Zhou$^{3}$ \quad Yuqian Xu$^{4}$ \quad Shuai Yang$^{1,*}$ \quad Xin Wu$^{1}$ \quad Zitong Zhou$^{1,5,\dagger}$ \quad Hui Liu$^{1}$ \\
  \textbf{Bin Zha}$^{1}$ \quad \textbf{Zheming Wang}$^{1}$ \quad \textbf{Liya Li}$^{1}$ \quad \textbf{Wei Wei}$^{1}$ \quad \textbf{Jinru Ding}$^{6}$ \quad \textbf{Wenrao Pang}$^{6}$ \\
  \textbf{Mouxiao Bian}$^{6}$ \quad \textbf{Haoyuan Hu}$^{1}$ \quad \textbf{Jun Xu}$^{1}$ \quad \textbf{Jie Xu}$^{6,*}$ \\[0.5em]
  {$^1$JD Health International Inc.} \\
  {$^2$Shanghai Jiao Tong University} \\
  {$^3$National University of Singapore} \\
  {$^4$University of North Carolina Chapel Hill} \\
  {$^5$University of Pennsylvania} \\
  {$^6$Shanghai Artificial Intelligence Laboratory}
}

\begin{document}
\maketitle
{\renewcommand{\thefootnote}{$\dagger$}\footnotetext{Work done during the internship at JD Health International Inc.}}
{\renewcommand{\thefootnote}{*}\footnotetext{Correspondence: Runhan Shi (\texttt{han.run.jiangming@sjtu.edu.cn}); Shuai Yang (\texttt{yangshuai15@jd.com}); Jie Xu (\texttt{xujie@pjlab.org.cn})}}

\begin{abstract}
Large language models (LLMs) are increasingly deployed in online medical consultation, yet existing benchmarks remain poorly aligned with real clinical practice. Many rely on synthetic conversations or patient simulators, omit patient-uploaded medical images, or evaluate open-ended clinical responses using multiple-choice or lexical-overlap metrics that poorly reflect clinical quality. We introduce \textbf{MedRealMM}, a large-scale benchmark for multimodal online medical consultation built from de-identified patient-doctor interactions collected from a nationwide Chinese internet hospital. MedRealMM uses a Multimodal Clinical Challenge Point (MCCP) extraction framework to identify clinically demanding moments in authentic consultation trajectories and converts each into a standardized next-response generation task while preserving the preceding text-image context. Each instance is paired with a case-specific rubric refined by physicians that rewards clinically desirable behaviors and penalizes unsafe, unsupported, or contradictory responses. The current release contains 5,620 real-world multimodal cases spanning 64 clinical departments. We evaluate 19 general-purpose and medical-specialized LLMs, including text-only and multimodal systems. Our results show that image information is critical for reliable clinical performance and that current frontier models remain below the online physician response. Although some frontier models satisfy as many or more positive clinical criteria than physicians, they trigger more negative criteria, indicating that safety-sensitive error avoidance remains a central bottleneck. MedRealMM offers a realistic and reproducible benchmark for evaluating multimodal medical reasoning in real-world online consultation. The dataset will be publicly available on Hugging Face at \url{https://huggingface.co/datasets/jdh-algo/MedRealMM}.
\end{abstract}


\section{Introduction}
\label{sec:intro}

Online medical consultation has become an important component of China's digital health system. Patients increasingly interact with physicians through asynchronous, platform-mediated encounters that combine free-text messages, structured medical records, and patient-uploaded images. This trend is driven by the uneven distribution of offline medical resources, the growing population of people with chronic diseases, and national digital health initiatives~\citep{jamies2026digital,gac2026telemedicine}. For example, JD Health (\url{https://ir.jdhealth.com}), one of the largest online healthcare platforms in China, reported 217.7 million annual active users at the end of 2025. Its Internet Hospital handled more than 180 million consultations in 2024, averaging more than 490,000 consultations per day~\citep{jdhealth2025annual,jdhealth2025llm}. The scale of these services has created a growing demand for AI systems that can assist physicians during online consultation.

\begin{figure*}[!t]
  \centering
  \includegraphics[width=0.95\textwidth]{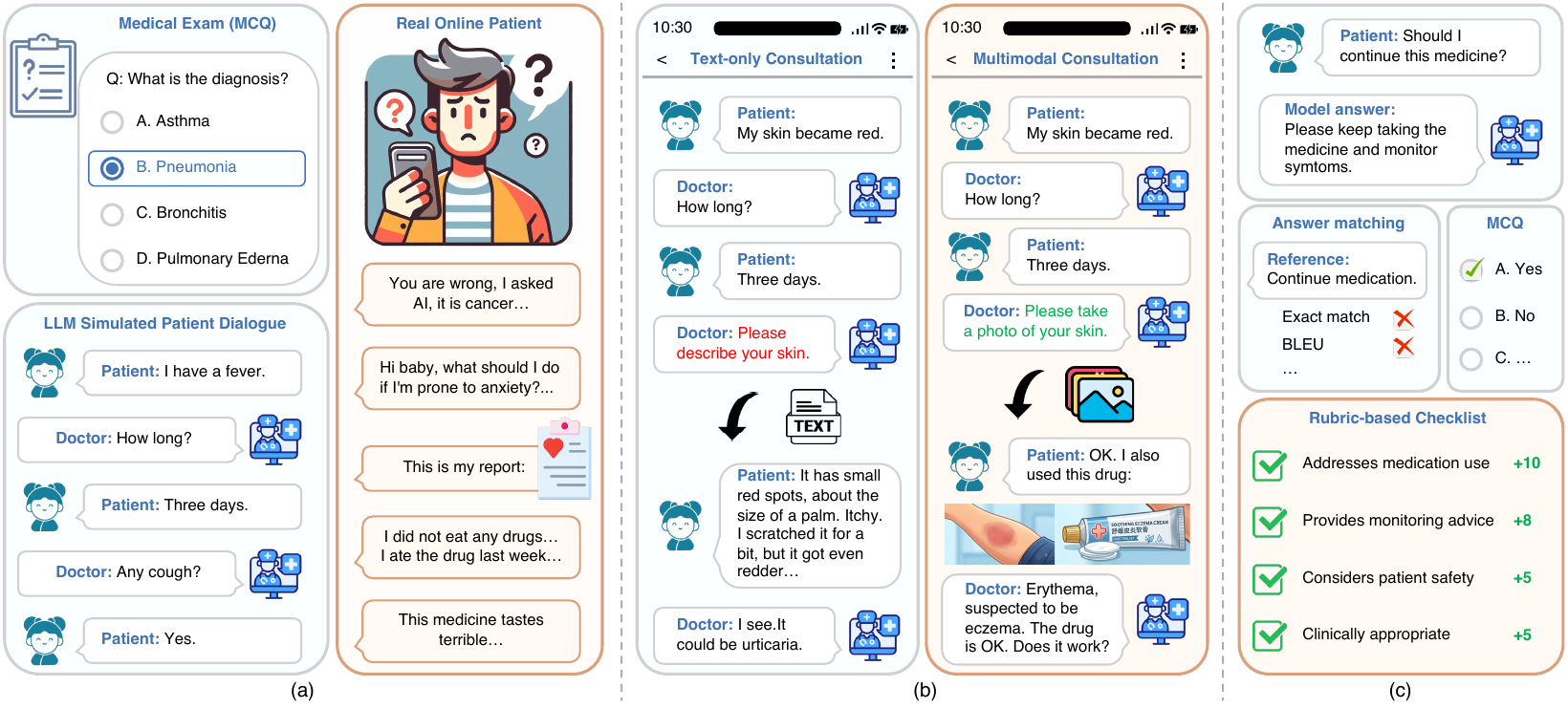}
  \caption{Three gaps between existing medical LLM benchmarks and real-world online consultation: (a) unrealistic consultation trajectories, (b) missing multimodal evidence, and (c) inadequate open-ended response evaluation.}
  \label{fig:limitations}
\end{figure*}

Recent advances in large language models (LLMs) have made them promising assistants for online medical consultation and have been increasingly explored for applications such as medical question answering, consultation assistance, and clinical decision support~\citep{Thirunavukarasu2023,Chen2024,McDuff2025}. However, whether these models can reliably support real-world online consultation remains largely unclear, since existing evaluations primarily test medical knowledge or simplified interactions rather than deployment-like consultation behavior.
As illustrated in Figure~\ref{fig:limitations}, this mismatch has three dimensions. First, many benchmarks use clean exam-style cases, short-form QA, synthetic dialogues, or patient simulators, while real patients often communicate in incomplete, colloquial, fragmented, and sometimes inconsistent ways across multiple turns. Second, medical dialogue benchmarks are typically text-only, while multimodal medical benchmarks usually evaluate curated image-question pairs; in contrast, online physicians must interpret patient-uploaded photographs, laboratory reports, prescriptions, or imaging results in the context of the evolving dialogue. Third, physician responses are open-ended, case-specific, and safety-sensitive, so reference matching, lexical overlap, and multiple-choice metrics fail to capture clinical appropriateness, missing information gathering, unsupported claims, or unsafe recommendations.

\begin{figure*}[t!]
  \centering
  \includegraphics[width=0.95\textwidth]{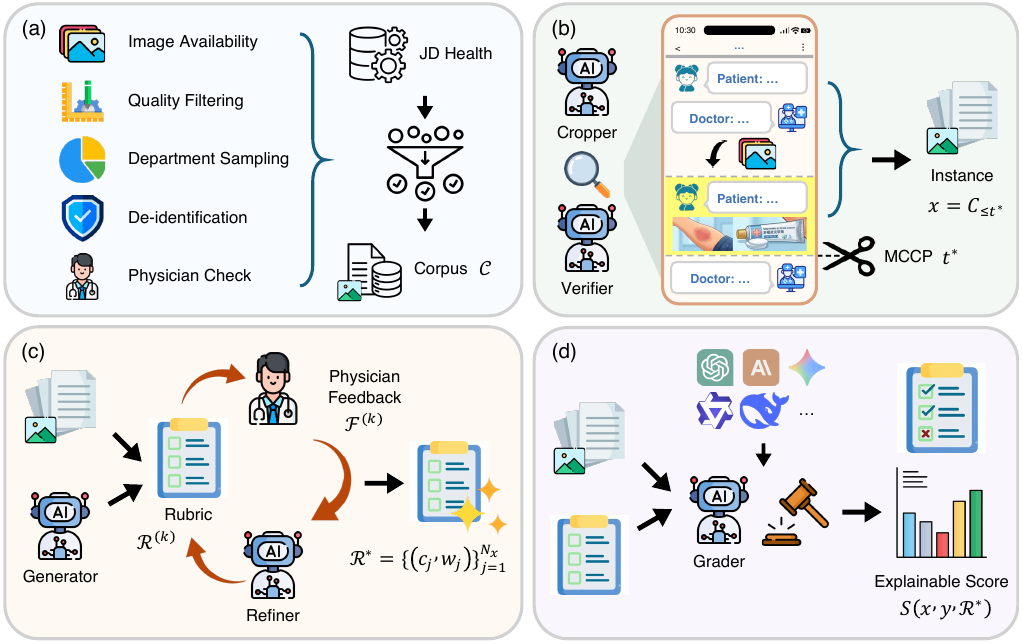}
  \caption{Overall \name construction pipeline. \textbf{(a) Corpus Construction}: multistage filtering and de-identification of real consultation logs from JD Health. \textbf{(b) MCCP Extraction}: an LLM agent identifies the clinically critical turn and constructs a self-contained benchmark instance. \textbf{(c) Rubric Construction}: iterative physician-guided refinement of a case-specific rubric. \textbf{(d) Automated Evaluation}: an LLM agent scores model responses against the approved rubric to produce final evaluation scores.}
  \label{fig:pipeline}
\end{figure*}

To address these gaps, we introduce \textbf{\name}, a real-world multimodal benchmark for Chinese online medical consultation constructed from de-identified one-on-one consultations collected from JD Health. Figure~\ref{fig:pipeline} provides an overview of the benchmark construction pipeline. Specifically, rather than replaying entire consultation trajectories or relying on patient simulation, we identify \emph{Multimodal Clinical Challenge Points} (MCCPs), namely clinically demanding points at which physicians must engage substantively, collecting information, interpreting images, performing differential diagnosis, recommending treatment, or communicating risk. Each MCCP is converted into a self-contained single-turn evaluation instance while preserving the original multimodal consultation context. The evaluation task is formulated as generating the physician's next response given the same multimodal context. This formulation evaluates multimodal understanding in its original consultation setting, where patient-uploaded images are embedded in multi-turn dialogue and must be interpreted jointly with the textual context. Each case is associated with a case-specific evaluation rubric, initially generated by an LLM and subsequently refined through iterative physician feedback. This process reduces the cost of annotation relative to fully manual rubric construction while maintaining physician oversight throughout rubric refinement.

Our contributions are summarized as follows: 
\begin{itemize}
    \item We introduce \name, a real-world multimodal benchmark for Chinese online medical consultation. Each benchmark case preserves the patient-uploaded medical images available during the original consultation.
    \item We propose \emph{Multimodal Clinical Challenge Point} (MCCP) extraction, a method that converts real multi-turn multimodal consultations into self-contained single-turn evaluation instances without resorting to patient simulation.
    \item We develop a physician-in-the-loop iterative rubric refinement protocol that combines LLM-initialized criteria with physician revisions, providing a case-specific evaluation grounded in real clinical judgment.
    \item We perform comprehensive evaluations of general-purpose, multimodal, and medical LLMs on \name, and find that top models match or exceed the treating physician on positive criteria yet trail on negatives, suggesting that patient safety remains an obstacle to deploying LLMs in real online consultation.
\end{itemize}

\section{Related Work}
\label{sec:related}

Medical LLM benchmarks differ in the realism of consultation scenarios, the use of multimodal medical information, and the evaluation of open-ended responses. Table~\ref{tab:benchmark-comparison} summarizes the relationship between the existing benchmarks and \name.

\subsection{Realism of Consultation}
\label{sec:related-realism}

Early medical LLM benchmarks primarily evaluate medical knowledge through question answering, including PubMedQA~\citep{pubmedqa2019}, MedQA~\citep{medqa2020}, MedMCQA~\citep{medmcqa2022}, and MultiMedQA~\citep{multimedqa2023,medpalm2_2025}. Chinese benchmarks similarly focus on medical examinations and clinical knowledge assessment~\citep{cblue2021,promptcblue2023,cmexam2023,cmb2023,medbench2024,medbenchv4_2025}. Although these resources measure medical knowledge, they do not evaluate the patient-physician interaction.
Subsequent work moves toward consultation settings. Dialogue datasets such as MedDialog, MedDG, and MediTOD~\citep{meddialog2020,meddg2020,meditod2024} provide real consultation transcripts but treat them as next-utterance prediction rather than performance evaluation. 
Recent benchmarks instead employ patient simulators or agent-based environments ~\citep{liao2023multiturn,mediq2024,agentclinic2024,maque2025,medconsultbench2026,meddialbench2026} and synthetic conversations~\citep{healthbench2025,meddialogrubrics2026} to evaluate interactive consultation. 
However, these benchmarks may not be able to reflect the distributional irregularities, such as fragmented and inconsistent behaviors, observed in real consultations.

The closest realistic settings include HealthBench Professional~\citep{healthbenchprofessional2026} and LiveMedBench~\citep{livemedbench2026}. HealthBench Professional focuses on evaluating clinician-facing tasks rather than online consultation. LiveMedBench is constructed from public medical forums but explicitly discards image-containing threads. In contrast, \name is built from private one-on-one patient-physician consultations in a deployed Chinese internet hospital and retains the patient-uploaded images that accompany them, reducing distribution mismatch and potential data contamination.

\subsection{Multimodal Medical Consultation}
\label{sec:related-multimodal}

Multimodal medical evaluation has largely focused on visual question answering over curated medical images, including PathVQA~\citep{pathvqa2020}, SLAKE~\citep{slake2021}, PMC-VQA~\citep{pmcvqa2023}, OmniMedVQA~\citep{omnimedvqa2024}, and GMAI-MMBench~\citep{gmaimmbench2024}. These benchmarks evaluate image understanding using isolated image-question pairs.

Conversely, existing medical dialogue benchmarks are predominantly text-only and either exclude image-containing consultations or do not collect visual information~\citep{meddialog2020,meddg2020,healthbench2025,healthbenchprofessional2026,livemedbench2026,medconsultbench2026,meddialogrubrics2026}. A few recent benchmarks combine consultation scenarios with visual information, including 3MDBench~\citep{3mdbench2025} and the multimodal track of MedBench v4~\citep{medbenchv4_2025}, but they rely on curated or simulated images. In contrast, \name evaluates physician response generation using patient-uploaded images embedded within real consultation histories.

\subsection{Evaluation of Open-Ended Responses}
\label{sec:related-rubric}

Because physician responses admit multiple clinically appropriate forms, exact-match and lexical-overlap metrics such as BLEU~\citep{papineni-etal-2002-bleu} and ROUGE~\citep{lin-2004-rouge} are often insufficient for evaluation. Recent work, therefore, adopts rubric-based evaluation using LLM judges.
HealthBench~\citep{healthbench2025} and HealthBench Professional~\citep{healthbenchprofessional2026} employ physician-written case-specific rubrics, providing clinically meaningful evaluation at a relatively high annotation cost. Subsequent approaches, including LiveMedBench and automated rubric pipelines~\citep{livemedbench2026,meddialogrubrics2026,automatedrubrics2026}, use LLM-generated rubrics with limited physician verification to improve scalability.

\name adopts a physician-guided rubric construction process in which case-specific rubrics are initialized by an LLM and iteratively refined by physicians. This design combines scalable rubric generation with continuous clinical supervision and supports the evaluation of real multimodal consultation cases.

\begin{table}[tbhp]
  \centering
  \small
  \setlength{\tabcolsep}{4pt}
  \renewcommand{\arraystretch}{1.2}
  \caption{Comparison of \name with the most related medical benchmarks. \textbf{Image}: \checkmark{} real patient-uploaded images form part of the input; $\times$ text-only; $\sim$ curated or simulated images; \textbf{Unit} denotes the evaluated interaction granularity; \textbf{Rubric} shows whether it adopts per-case criteria for open-ended responses.}
  \label{tab:benchmark-comparison}
  \begin{tabular}{llccccc}
    \toprule
    \textbf{Benchmark} & \textbf{Data source} & \textbf{Size} & \textbf{Language} & \textbf{Image} & \textbf{Unit} & \textbf{Rubric} \\
    \midrule
    MedDialog-CN (2020)     & Real consultation & 1.1M & ZH & $\times$           & Multi-turn   & $\times$ \\
    3MDBench (2025)         & Patient/doctor simulator    & 3K       & EN    & $\sim$             & Multi-turn   & $\times$ \\
    HealthBench (2025)      & Synthetic     & 5K      & Multi & $\times$           & Single-turn   & \checkmark \\
    MedBench v4 (2025)      & Exam + mixed    & 700K+    & ZH    & $\sim$             & Mixed        & $\times$ \\
    HealthBench Pro (2026)  & Real clinician  & 525    & EN    & $\times$           & Single-turn & \checkmark \\
    MedConsultBench (2026)  & Patient simulator & 35.8K  & EN    & $\times$           & Multi-turn   & $\times$ \\
    MedDialogRubrics (2026) & Patient simulator  & 5.2K  & EN    & $\times$           & Multi-turn   & \checkmark \\
    LiveMedBench (2026)     & Real forum Q\&A  & 2.8K   & Multi & $\times$ & Single-turn       & \checkmark \\
    \midrule
    \textbf{\name (Ours)}   & \textbf{Real consultation} & 5.6K & \textbf{ZH} & \checkmark & \textbf{Single-turn} & \checkmark \\
    \bottomrule
  \end{tabular}
\end{table}

\section{Methods}
\label{sec:methods}

\subsection{Phase 1: Construction of a Real-World Multimodal Consultation Corpus}
\label{sec:data-processing}

\paragraph{Data source.}
MedRealMM is constructed from real patient-doctor interactions collected from the large-scale Chinese internet hospital platform JD Health between 2025 and 2026~\citep{jdhealth2025annual,jdhealth2025llm}. The platform serves patients nationwide across a broad spectrum of clinical departments and supports multimodal consultations, where both patients and physicians can upload medical images alongside textual descriptions of their symptoms.

Formally, each consultation is represented as an ordered multimodal interaction trajectory
\begin{align}
C=(e_1,e_2,\ldots,e_T),\ e_t=(c_t,m_t,s_t,\tau_t),
\end{align}
where $c_t$ denotes the content of the event $e_t$, $m_t\subseteq\{\text{text},\text{image}\}$ its modality set (a single event can carry both text and image attachments), $s_t\in\{\text{patient},\text{physician}\}$ the sender identity, and $\tau_t$ the timestamp. This formulation preserves both the temporal ordering of events and the interleaving of textual and visual information.

\paragraph{Multistage corpus construction.}
Starting from the raw record set $\mathcal{C}_0$, we obtain the final corpus $\mathcal{C}$ through a chain of filtering, balancing, and review operators
\begin{equation}
    \mathcal{C}_0 \xrightarrow{\,\Phi_{\text{img}}\,}
    \mathcal{C}_1 \xrightarrow{\,\Phi_{\text{cmpl}}\,}
    \mathcal{C}_2 \xrightarrow{\,\Phi_{\text{bal}}\,}
    \mathcal{C}_3 \xrightarrow{\,\Phi_{\text{deid}}\,}
    \mathcal{C}_4 \xrightarrow{\,\Phi_{\text{rev}}\,}
    \mathcal{C},
\end{equation}
where each operator performs one stage of corpus construction.
\textbf{(i) Image presence}: $\Phi_{\text{img}}$ retains only consultations containing at least one patient-uploaded medical image, $\mathcal{C}_1=\{C\in\mathcal{C}_0\mid \exists\,e_t\in C,\,\text{image}\in m_t\}$, ensuring that every retained consultation contains patient-provided visual evidence.
\textbf{(ii) Completeness}: $\Phi_{\text{cmpl}}$ removes consultations that are excessively short, truncated, or otherwise unsuitable for meaningful evaluation.
\textbf{(iii) Department balance}: $\Phi_{\text{bal}}$ performs department-aware sampling to alleviate the severe specialty imbalance in the raw data.
\textbf{(iv) De-identification}: $\Phi_{\text{deid}}$ applies an LLM-assisted de-identification~\citep{liu2025deidgptzeroshotmedicaltext} that removes personally identifiable information from both text and images while preserving clinically relevant content; cases whose clinical interpretation depends primarily on identifiable visual information are excluded.
\textbf{(v) Manual review}: $\Phi_{\text{rev}}$ requires trained reviewers to verify both privacy protection and preservation of clinical utility.
The final corpus
\begin{equation}
\mathcal{C}=\{C_1,C_2,\ldots,C_N\}
\end{equation}
is used in the subsequent stages of benchmark construction. Detailed filtering rules, de-identification procedures, and privacy considerations are provided in Appendix~\ref{app:filtering-pipeline}.

\subsection{Phase 2: Multimodal Clinical Challenge Point Extraction}
\label{sec:mccp}

A realistic evaluation of multi-turn online consultation would ideally replay the entire interaction. In practice, this requires patient simulation, which often fails to capture the complexity and unpredictability of real patient behavior~\citep{meddialogrubrics2026,medconsultbench2026}. We instead identify clinically meaningful points within authentic consultation trajectories and formulate evaluation around these states. This leads to the notion of a Multimodal Clinical Challenge Point (MCCP).

\paragraph{Definition of MCCP.}
Given a consultation trajectory $C=(e_1,e_2,\ldots,e_T)$, an MCCP corresponds to a position $t^* \in \{1,\ldots,T\}$ immediately before a physician response, at which the accumulated consultation state requires substantive clinical reasoning based on the clinical expertise of the responding physician. A valid MCCP must jointly satisfy the following:

(i) \textbf{Clinical engagement} $\mathcal{A}(t^{*}; C)=1$: the expected physician response involves substantive clinical engagement (e.g., history taking, differential diagnosis, examination or imaging interpretation, risk stratification, treatment planning, or patient education) rather than routine acknowledgments or administrative replies.

(ii) \textbf{Multimodal relevance} $\mathcal{M}(t^{*}; C)=1$: images uploaded in $C_{\le t^{*}}$ can materially influence an appropriate physician response, including diagnostic reasoning, risk assessment, or treatment decisions.

A position $t^{*}$ is admitted as an MCCP iff $\mathcal{A}(t^{*};C)=1\wedge\mathcal{M}(t^{*};C)=1$.

\paragraph{MCCP extraction pipeline.}
We implement the two extraction criteria through a two-agent pipeline.
Let $\mathcal{T}=\{t\;|\;e_{t+1} \text{is a physician response}\}$ denote the set of turns before a physician response in the consultation trajectory $C$. Each agent $\phi:(t,C)\rightarrow(l,s)$ takes a trajectory and a candidate turn as input, and returns a binary decision $l\in\{0,1\}$ together with a selection score $s\in\mathbb{R}$. Binary decisions correspond to the clinical engagement criterion $\mathcal{A}$ and the multimodal relevance criterion $\mathcal{M}$, respectively, while selection scores are used to rank multiple valid candidates.

\textbf{Stage 1: Filtering for clinical engagement.}
Given the complete consultation trajectory $C$, an LLM agent \textit{Cropper} $\phi_\mathcal{A}$ evaluates each physician response turn and retains turns that satisfy the clinical engagement criterion:
\begin{equation}
(l_\mathcal{A}(t),\,s_\mathcal{A}(t))=\phi_{\mathcal{A}}(t,C),\quad \mathcal{T}_{\mathcal A}(C)=\left\{t\in\mathcal{T}\;\middle|\;l_\mathcal{A}(t)=1\right\}.
\end{equation}

\textbf{Stage 2: Verification of multimodal relevance.}
For each candidate turn, an LLM agent \textit{Verifier} $\phi_\mathcal{M}$ evaluates the consultation history available until the physician response and retains turns whose accompanying patient images are clinically necessary to
determine an appropriate response:
\begin{equation}
(l_\mathcal{M}(t),\,s_\mathcal{M}(t))=\phi_\mathcal{M}(t, C_{\le t}), \quad \mathcal{T}_{M}(C)=\left\{t\in\mathcal{T}_{\mathcal A}(C)\;\middle|\;l_\mathcal{M}(t)=1\right\}.
\end{equation}

If multiple candidate turns remain, we select the MCCP as $t^{*}=\arg\max_{t\in\mathcal{T}_{M}(C)}\left(s_\mathcal{A}(t)+s_\mathcal{M}(t)\right)$. Consultations with $\mathcal{T}_\mathcal{M}(C)=\emptyset$ are discarded. The extracted MCCPs are independently validated by physician review (Appendix~\ref{app:mccp-validation}).

\paragraph{Benchmark instance construction.}
For each retained consultation, the trajectory is truncated at its MCCP, and the benchmark instance is defined as
\begin{equation}
    x = C_{\le t^{*}}=\{e_1,e_2,\ldots,e_{t^{*}}\},
\end{equation}
where $e_{t^{*}}$ denotes the patient turn, and $x$ contains all textual and visual information available up to and including this patient turn. The prediction target is the turn of the physician immediately following $y_{\text{target}}=c_{t^{*}+1}$. The evaluation task is to generate $\hat{y}\sim p_\theta(\cdot\mid x)$ conditioned on this truncated multimodal state. This formulation transforms raw consultation logs into self-contained evaluation instances while yielding standardized evaluation instances derived directly from real consultation trajectories.

\paragraph{Patient-intent and consultation-stage annotation.}
Online consultations vary substantially in both patient intents and consultation stages. Reporting only aggregate performance obscures these differences and makes model failures difficult to interpret. To enable a stratified and clinically interpretable evaluation, we further annotate each instance $x$ with two structured labels:
\begin{equation}
  z_{\mathrm{intent}}=\phi_{\mathcal{I}}(x)\in\mathcal{I},\qquad z_{\mathrm{stage}}=\phi_{\mathcal{S}}(x)\in\mathcal{S},
\end{equation}
where $\phi_{\mathcal{I}}$ is an LLM agent that predicts the dominant patient intent of a closed taxonomy $\mathcal{I}$ (e.g., diagnosis seeking, examination interpretation, treatment consultation, medication guidance, follow-up), and $\phi_{\mathcal{S}}$ predicts the consultation stage of a clinically motivated taxonomy $\mathcal{S}$ aligned with standard medical-consultation practice (e.g., history taking, examination/image review, diagnostic reasoning, management planning). 
These annotations enable stratified evaluation, support balanced benchmark construction, and facilitate downstream error analysis.

\subsection{Phase 3: Physician-Guided Case-specific Rubric Construction}
\label{sec:rubric}
Evaluating open-ended medical consultation responses cannot be reliably reduced to reference-answer matching~\citep{healthbench2025,livemedbench2026}, as multiple clinically appropriate responses may exist for the same consultation state. We therefore construct a physician-guided, case-specific rubric for each benchmark instance.

\paragraph{Case-specific rubric generation.}
Given a benchmark instance $x$ and the corresponding original physician response $y_{\text{orig}}$, an LLM agent \textit{Generator} $\phi_{\mathcal{R}}$ generates an initial rubric
\begin{equation}
\mathcal{R}^{(0)}=\phi_{\mathcal{R}}(x, y_{\text{orig}})=\{(c_j,w_j)\}_{j=1}^{N_x},
\end{equation}
where each criterion $c_j$ is a clinically meaningful evaluation criterion that should be satisfied or violated by an appropriate physician response, and $w_j\in[-20,20]\setminus\{0\}$ denotes its signed importance. We divide the rubric into a \emph{positive set} $\mathcal{R}^{+}=\{j:w_j>0\}$ that rewards the inclusion of clinically desirable behaviors (e.g., correct differential, examination ordering, and safety advice), and a \emph{negative set} $\mathcal{R}^{-}=\{j:w_j<0\}$ that penalizes hallucinations, contradictions, or unsafe behaviors. The rubric size $N_x$ is instance-dependent and adapts to the clinical complexity of each case.

\paragraph{Iterative rubric refinement.}
In each round $k$, physicians review $\mathcal{R}^{(k)}$ and provide structured feedback $\mathcal{F}^{(k)}$ identifying missing criteria, incorrect weights, or overlooked clinical considerations, such as unsupported diagnoses, misinterpretations of images, unsafe recommendations, or missing follow-up advice. A rubric-improvement \textit{Refiner} agent $\phi_{\mathcal{R}}^{\,\prime}$ then revises the rubric:
\begin{equation}
\mathcal{R}^{(k+1)}=\phi_{\mathcal{R}}^{\,\prime}\!\left(x,\mathcal{R}^{(k)},\mathcal{F}^{(k)}\right).
\end{equation}
The procedure ends in round $K=\min\{k:\mathcal{F}^{(k)}=\varnothing\}$, i.e., when the physicians approve the rubric without further revisions. Because JD Health imposes a strict per-turn length limit to encourage concise physician responses, we incorporate case-specific length constraints as rule-based evaluation criteria. We denote the resulting physician-approved rubric by $\mathcal{R}^{*}=\mathcal{R}^{(K)}$ and use it for all downstream evaluations.

The original physician response is used only to initialize case-specific evaluation criteria rather than to define a reference answer. Through iterative physician refinement, criteria that are overly specific to the original response are revised or removed, resulting in a rubric that captures clinically acceptable behaviors instead of reference-specific wording.

\subsection{Phase 4: Rubric-Grounded Automated Evaluation}
\label{sec:grade}
MedRealMM evaluates open-ended medical consultation responses through rubric-grounded clinical assessment~\citep{healthbench2025}:
\begin{align}
\mathrm{Grade}(x,y,\mathcal{R}^{*})\rightarrow S,
\end{align}
where $x$ denotes the benchmark instance, $y$ denotes the generated physician response, $\mathcal{R}^{*}$ denotes the physician-approved case-specific rubric, and $S$ denotes the resulting evaluation score. 

\paragraph{Evaluation setup.}
For each benchmark instance $x=C_{\le t^{*}}$, an evaluated model $M$ generates a physician response $y\sim p_{M}(\cdot\mid x)$, which is then assessed against the physician-approved rubric $\mathcal{R}^{*}=\{(c_j,w_j)\}_{j=1}^{N_x}$ obtained in Phase~\ref{sec:rubric}.

\paragraph{Criterion-level assessment.}
For each criterion $c_j$, an LLM agent \textit{Grader} $\phi_{\mathcal{J}}$ (LLM-as-a-judge~\citep{zheng2023judgingllmasajudgemtbenchchatbot,healthbench2025}) predicts a binary decision together with a rationale:
\begin{align}
(v_j, r_j)=\phi_{\mathcal{J}}\left(x,y,c_j\right),\qquad v_j\in\{0,1\},
\end{align}
where $v_j$ measures whether $y$ satisfies $c_j$ given the instance $x$, and $r_j$ is a textual rationale that explains why. $\phi_{\mathcal{J}}$ does not condition on the original physician response $y_{\text{orig}}$, so the generated answers are not penalized for stylistic deviation from the single observed reference.

\paragraph{Rubric-level aggregation.}
Criterion-level scores are aggregated according to their signed importance weights. Since negative weights can make $\sum_j w_j$ non-positive, we normalize only by the total positive weight and clip the resulting score to $[0,1]$:
\begin{align}
S(x,y,\mathcal{R}^{*})=\text{Clip}\left(\frac{\sum_{j=1}^{N_x}w_j v_j}{\sum_{j=1}^{N_x}\max(w_j, 0)}, 0, 1\right).
\end{align}

\paragraph{Benchmark-level evaluation.}
For an evaluated model $M$, the final benchmark score is the case-level mean over the evaluation set $\mathcal{D}$:
\begin{align}
\mathrm{Score}(M)=\frac{1}{|\mathcal{D}|}\sum_{i=1}^{|\mathcal{D}|}S\!\left(x_i,\,M(x_i),\,\mathcal{R}_i^{*}\right),
\end{align}
where $x_i\in\mathcal{D}$. 
This rubric-grounded evaluation framework enables scalable assessment of multimodal medical consultation responses while preserving case-specific clinical expectations and multimodal reasoning requirements.

\subsection{Dataset Statistics}
\begin{figure*}[!t]
  \centering
  \includegraphics[width=0.95\textwidth]{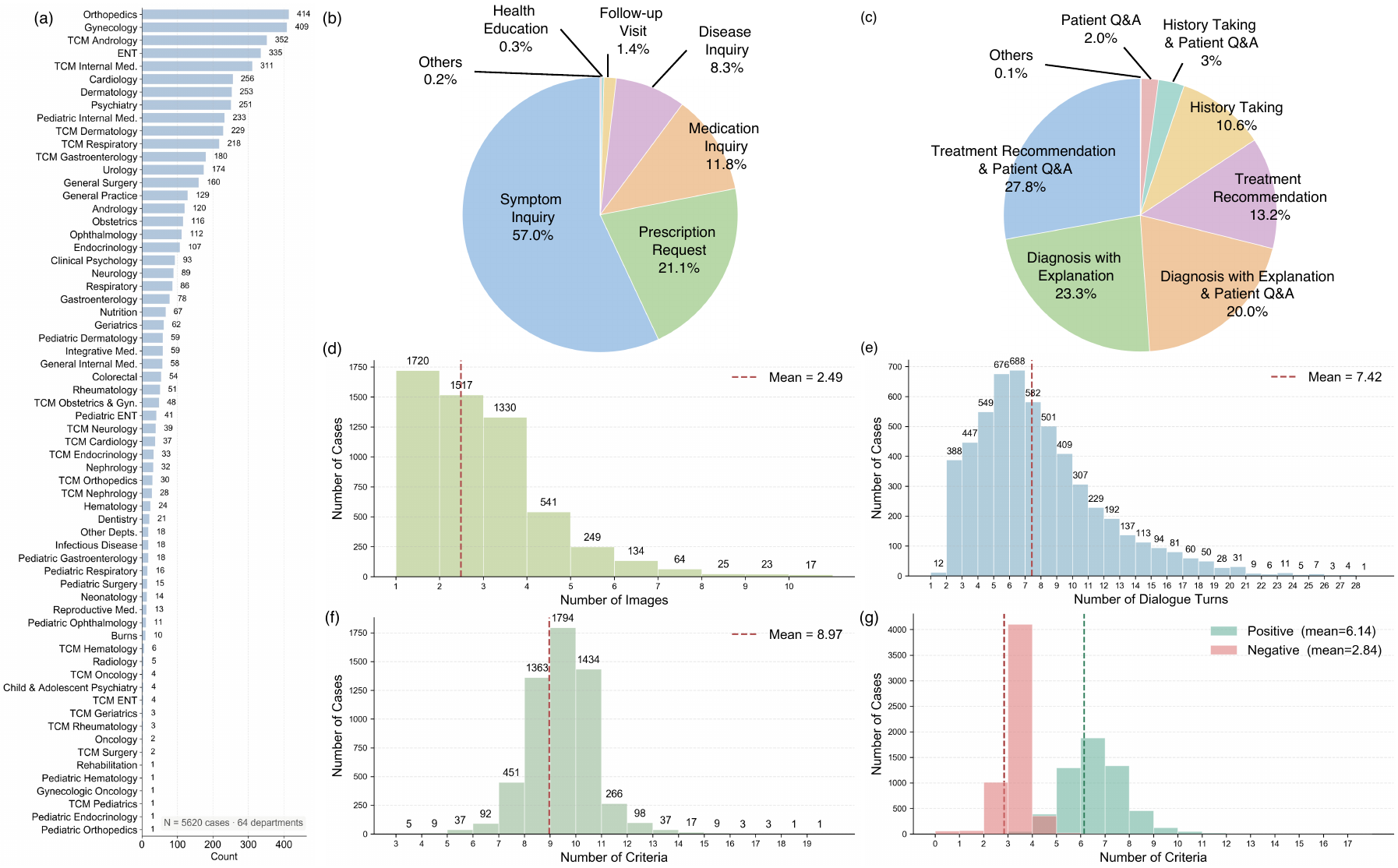}
  \caption{Data statistics of \name: distributions of (a) clinical departments, (b) patient intents, (c) consultation stages, (d) uploaded images per case, (e) dialogue turns per case, (f) rubric criteria per case, and (g) the split into positive and negative sets.}
  \label{fig:statistics}
\end{figure*}

\name contains 5,620 consultations spanning 64 clinical departments. Figure~\ref{fig:statistics} further summarizes the distributions of patient intents, consultation stages, dialogue turns, image counts, and rubric complexity. Symptom inquiry is the most common patient intent, while treatment recommendation and patient Q\&A account for the largest fraction of extracted MCCPs. 
The cases contain between 1 and 10 patient-uploaded images and an average of approximately 7 dialogue turns. The physician-in-the-loop rubric construction process yields 3 to 19 criteria per case (positive: 2 to 17, negative: 0 to 7), enabling fine-grained assessment of model responses.

\section{Experiments}
\label{sec:exp}

We first describe the experimental setup (Section~\ref{sec:settings}) and summarize the main results (Section~\ref{sec:main-results}). We then investigate three questions corresponding to the key contributions of \name. Section~\ref{sec:rq1} asks whether real multimodal consultation constitutes a distinct, unsolved capability, and tests the hypothesis that visual grounding rather than biomedical knowledge is the dominant bottleneck. Section~\ref{sec:rq2} examines where the models break down along the clinical axes that matter for safe remote care. Section~\ref{sec:rq3} assesses whether rubric-grounded LLM-as-a-judge scoring on real consultations is reliable enough to support these conclusions.

\subsection{Experiment Settings}
\label{sec:settings}

We evaluate 19 models as revealed in Table~\ref{tab:model-list}, spanning three dimensions: modality (multimodal vs.\ text-only), domain focus (general-purpose vs.\ medical-specialized), and access type (open-source vs.\ closed-source). 
Unless otherwise specified, Claude-Opus-4.7~\citep{claude4opus_versions} is used for rubric generation and refinement, while Gemini-3-Pro-Preview~\citep{geminiteam2025geminifamilyhighlycapable} serves as the \textit{Grader} and de-identification agent. Rubric refinement is performed for three iterations to achieve convergence. 
\name evaluates clinical quality but not response latency; in deployment, physicians are expected to respond within roughly 20 seconds, and jointly measuring quality and latency is beyond the scope of this benchmark. See Appendix~\ref{app:eval} for more details.

\subsection{Main Results}
\label{sec:main-results}

\begin{figure*}[tbhp]
  \centering
  \includegraphics[width=0.95\textwidth]{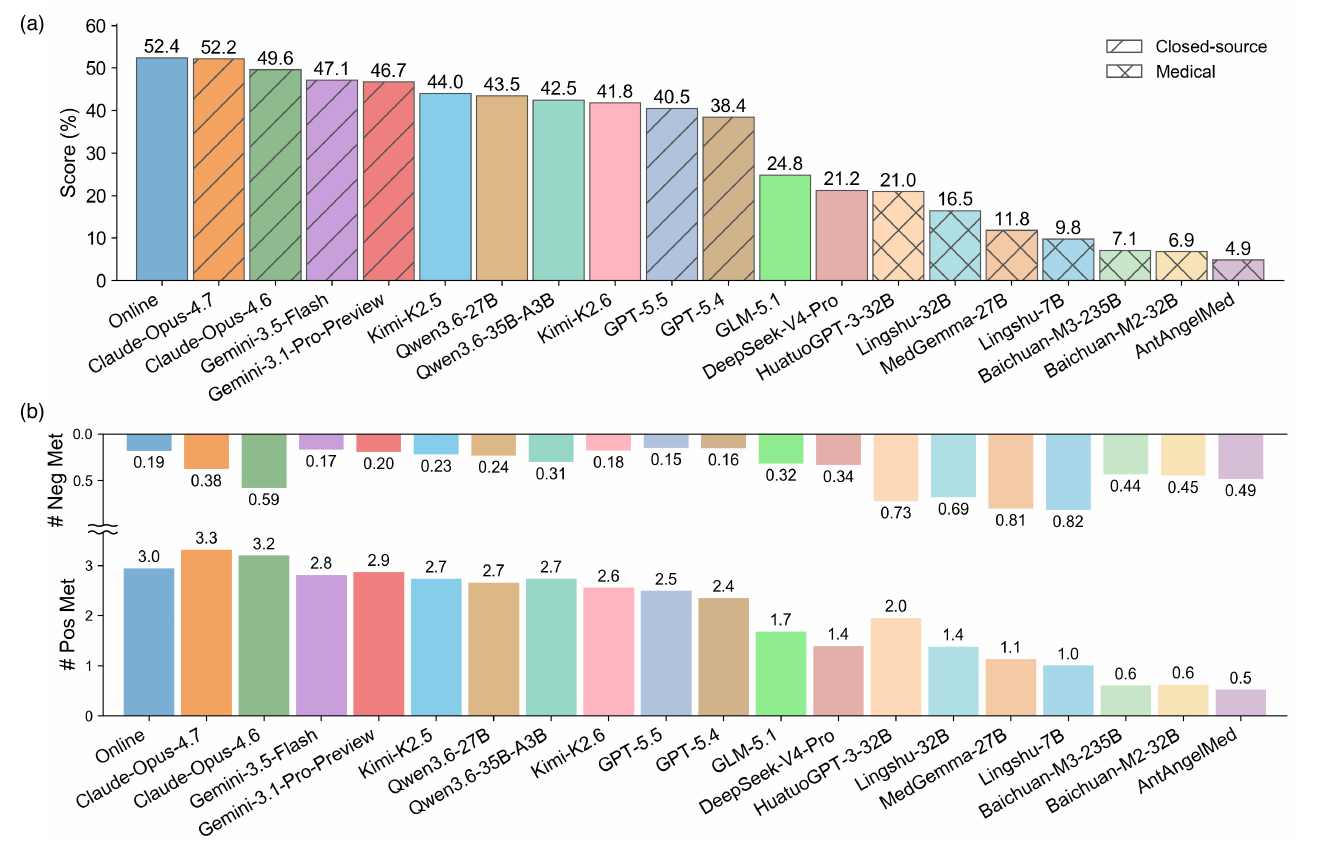}
  \caption{Evaluation results on \name (best available modality per model), including (a) overall performance and (b) per-case decomposition into positive (reward) and negative (penalty) criteria.}
  \label{fig:res_overall}
\end{figure*}

Figure~\ref{fig:res_overall} (a) summarizes the overall performance of all models, together with the original real-world physician response (\texttt{Online}). The ordering is consistent: a closed-source frontier cluster leads general-purpose open-source models, which in turn lead medical-specialized models, and the physician response sits above all of them. Even the best model falls short of \texttt{Online}, and its absolute scores remain around $50$, well below saturation, despite being at the frontier. 
Unlike prior medical benchmarks, every evaluation instance in \name is derived from an authentic consultation requiring joint reasoning over Chinese patient dialogue and uploaded medical images. Consequently, these scores better reflect deployment readiness than performance on exam-style medical benchmarks. See Section~\ref{sec:more-res} for additional results and analysis.

\paragraph{The physician anchor leads by avoiding penalties, not by covering more positives.}
Figure~\ref{fig:res_overall} (b) decomposes each model's score into the average number of positive (reward) and negative (penalty) criteria its responses satisfy per case. The two components become decoupled among the highest-performing systems. Claude-Opus-4.6 and 4.7 actually meet more positive criteria than the \texttt{Online} physician response, but \texttt{Online} ranks above them because it triggers fewer negatives. Below the top, the joint pattern is simpler: lower-scoring models meet fewer positives and trigger more negatives. In a clinical setting, a single harmful recommendation can outweigh several correct ones~\citep{hager2024evaluation}. These results suggest that avoiding clinically unsafe behavior, rather than maximizing positive coverage, is the key factor separating the treating physician from current frontier LLMs.

\subsection{Is Visual Grounding the Main Bottleneck?}
\label{sec:rq1}

\begin{figure*}[tbhp]
  \centering
  \includegraphics[width=0.95\textwidth]{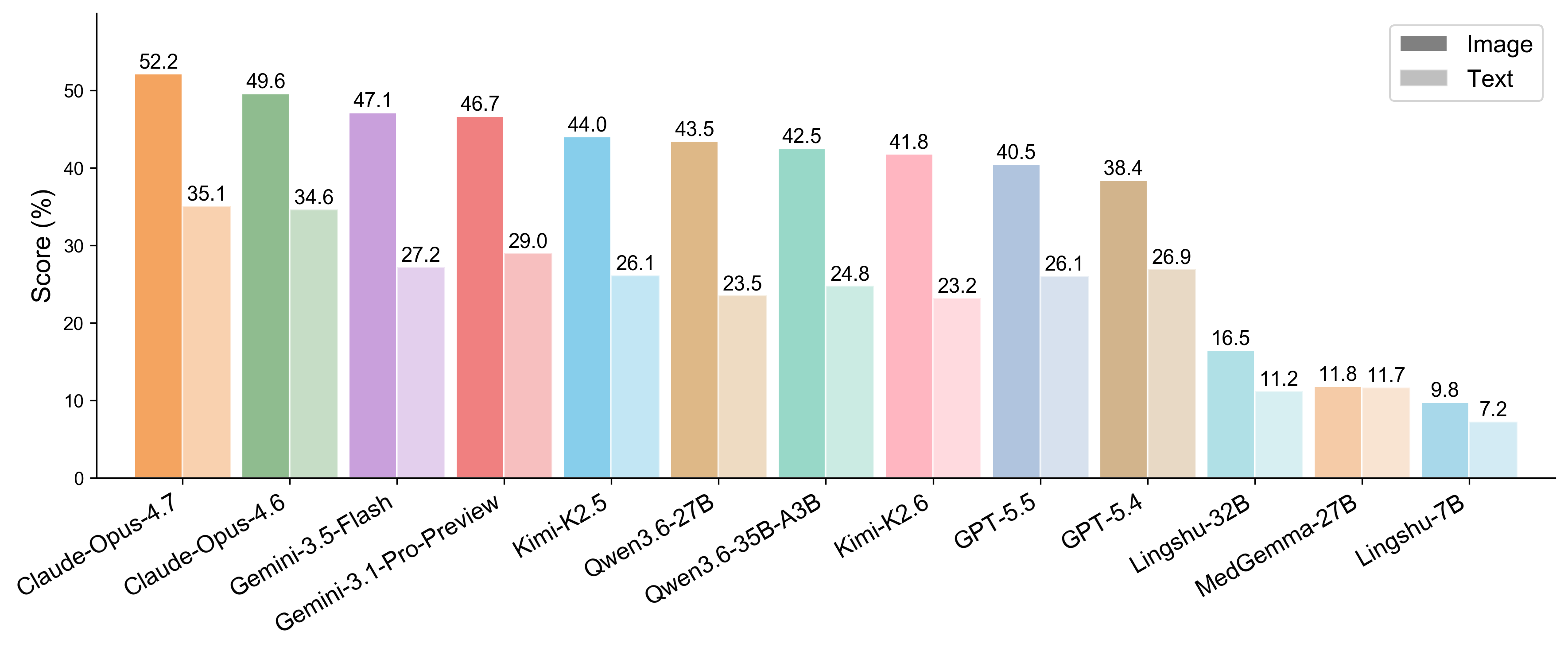}
  \caption{Multimodal (text\,+\,image) vs.\ text-only performance for multimodal models on \name. Darker and lighter bars denote multimodal input and text-only input, respectively.}
  \label{fig:res_image_text}
\end{figure*}

\paragraph{The closed-source advantage emerges primarily from multimodal capability.}
Figure~\ref{fig:res_image_text} compares the score of each multimodal model with and without patient images. Providing images raises frontier scores by 14 to 20 points, in some cases nearly doubling the text-only score. This behavior is consistent with the benchmark design: each MCCP is selected so that patient-uploaded images are clinically relevant to the subsequent physician response. On text-only input, the best open model (GLM-5.1 at 24.8) sits inside the frontier models' own text-only band of 23.2 to 35.1; the closed-source advantage widens substantially once images are available. Because nearly all existing medical dialogue benchmarks are text-only, they would compress these systems into a much narrower performance range and obscure the multimodal capability required in real consultation.

\paragraph{Medical-specialized models trail general-purpose ones and gain little from images.}
All medical-specialized models in our pool rank at the bottom. This trend is consistent with observations on the text-only LiveMedBench~\citep{livemedbench2026}; it persists and widens in the multimodal setting. This result likely reflects two factors. First, most medical-specialized models are text-only, so a multimodal benchmark places them at a structural disadvantage. Second, for the few that do accept images, the performance gain from image input is less than 6 points, substantially smaller than for general-purpose frontier models. These results suggest that current medical-specialized models are optimized primarily for text-based medical reasoning and derive limited benefit from visual evidence.

\subsection{Where do Models Fail?}
\label{sec:rq2}

\begin{figure*}[tbhp]
  \centering
  \includegraphics[width=0.95\textwidth]{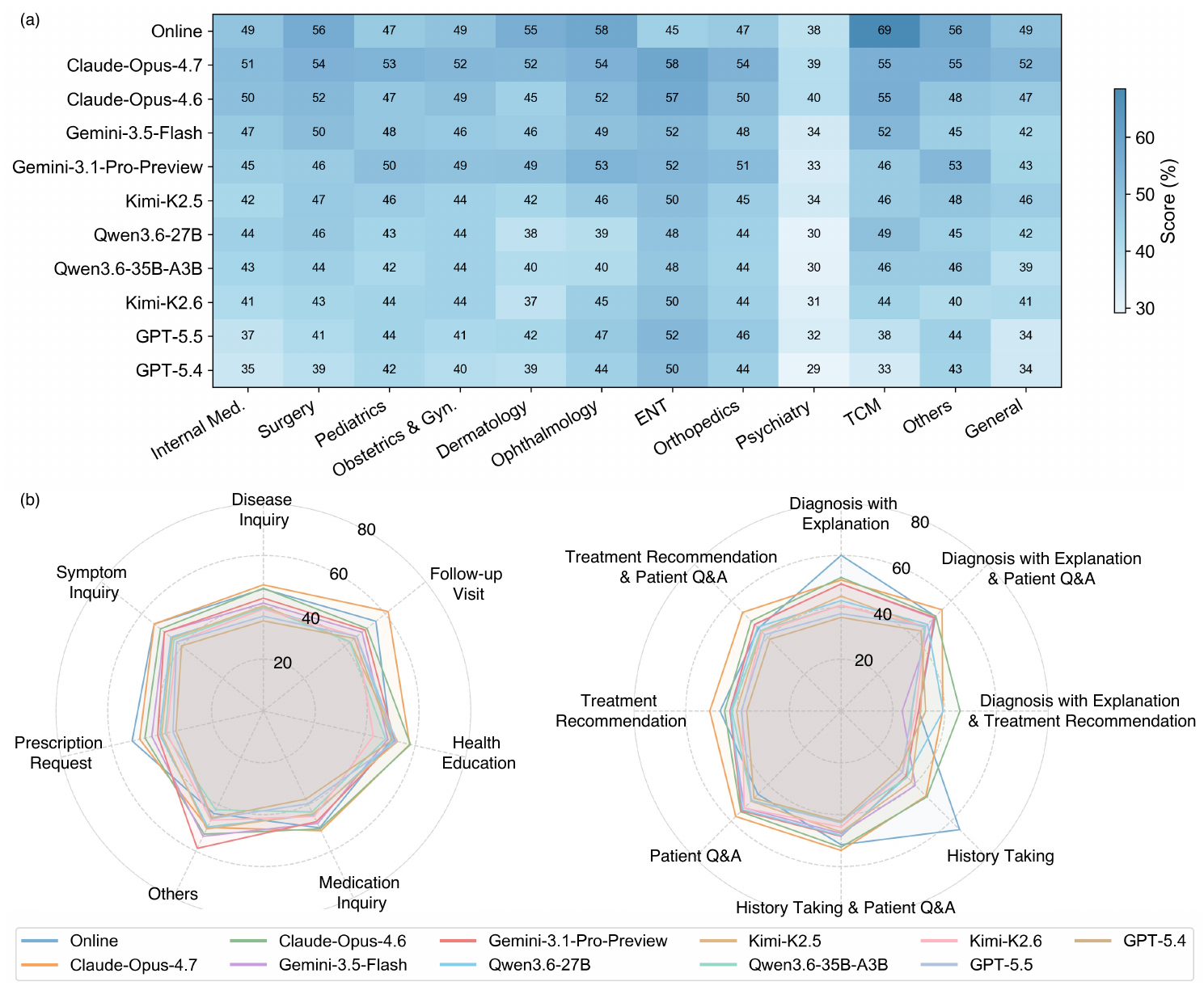}
  \caption{Results for representative models grouped by (a) major clinical departments, and (b) patient intents and consultation stages.}
  \label{fig:res_2}
\end{figure*}

\paragraph{Specialty robustness masks a Chinese-specific coverage gap.}
Figure~\ref{fig:res_2} (a) reports performance after aggregating the 64 clinical departments into major specialty groups. 
Frontier models show relatively consistent performance across most clinical specialties. For example, Claude-Opus-4.7 achieves scores between 51 and 58 in nearly all departments, suggesting that variation across models is generally larger than variation across specialties. Psychiatry is a notable exception: all evaluated systems, including \texttt{Online}, score less than or equal to 40, indicating that psychiatric consultation remains a challenging setting under our evaluation protocol for both current LLMs and treating physicians. 
traditional Chinese medicine (TCM) exhibits a different pattern. While frontier models achieve performance comparable to their results in other specialties, the treating physician outperforms the strongest LLM by 14 points, representing the largest \texttt{Online}-LLM gap in the benchmark. This suggests that the gap between current multimodal LLMs and practicing physicians is particularly pronounced in TCM, underscoring the importance of evaluating models on authentic Chinese consultation data rather than benchmarks centered on Western medical practice.

\paragraph{Failures are uniform across patient intents but concentrate at specific consultation stages.}
We further analyze model performance across patient intents and consultation stages, as shown in Figure~\ref{fig:res_2}(b). Across the seven intent classes, the frontier models hold a narrow 40 to 60 band, so the type of patient goal alone does not determine difficulty. However, two consultation stages consistently emerge as the most difficult. Most models score below 40 on the combined diagnosis-with-explanation and treatment-recommendation stage, where multiple clinical decisions should be reasoned about together. The clearest gap to the physician response appears in history taking: \texttt{Online} reaches roughly 70, while every model falls below 50. Success at this stage typically requires \emph{asking} for the missing information rather than answering immediately, the ``know-when-to-ask'' competence emphasized in interactive-consultation work~\citep{mediq2024,maque2025}.

\begin{figure*}[tbhp]
  \centering
  \includegraphics[width=0.95\textwidth]{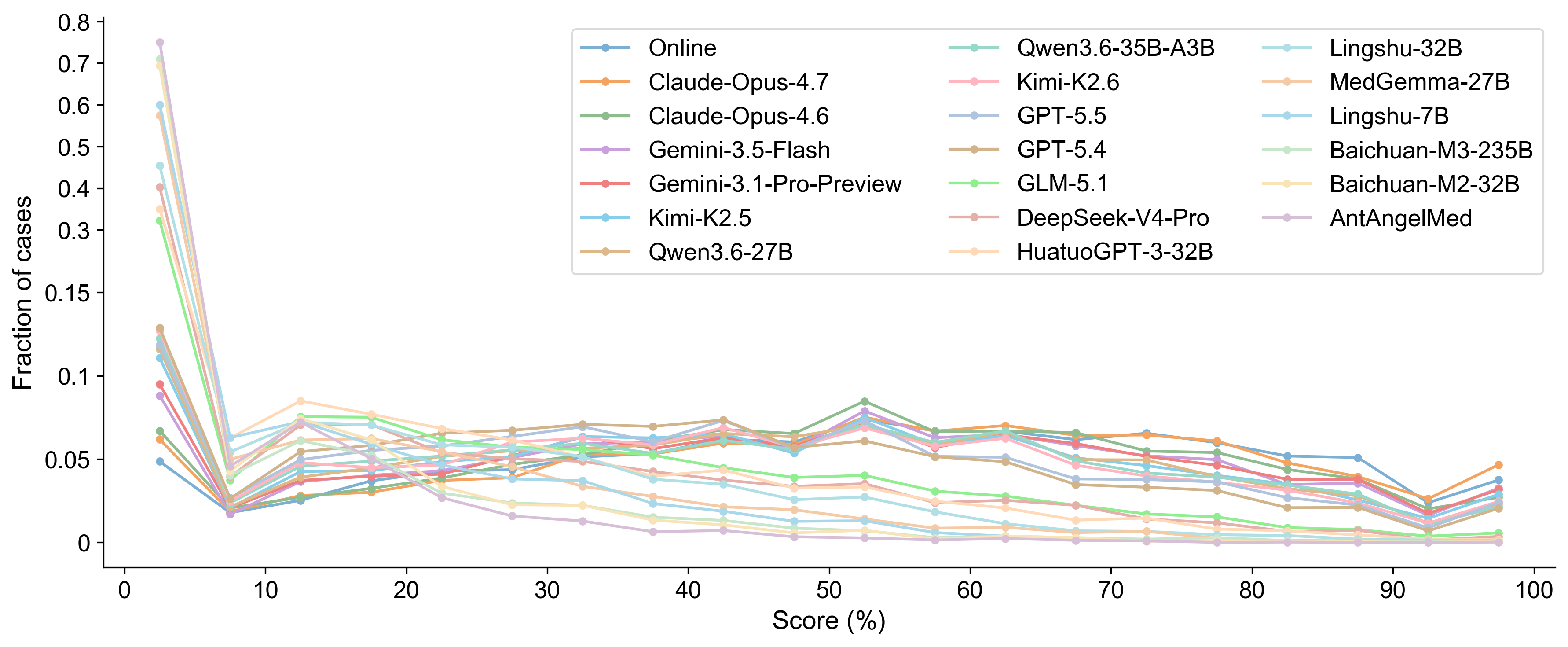}
  \caption{Per-case score distribution for evaluated models and \texttt{Online} physician response (bin width 5\%).}
  \label{fig:score_distribution}
\end{figure*}

\paragraph{\name cleanly separates models and has no ceiling effect.}
Figure~\ref{fig:score_distribution} reports the per-case score distribution. Weak models are unimodal with a large spike at 0: a response that meets too few positive criteria or triggers too many negative ones is assigned the minimum score, regardless of whether it is broadly wrong or only narrowly inadequate. Frontier models, in contrast, spread broadly across the range with negligible mass at either tail. This contrasts with HealthBench~\citep{healthbench2025}, whose per-case distribution exhibits two peaks: a larger one near 0 and a smaller one near the ceiling. The high-end peak is absent from \name (even the strongest systems rarely score near 100 on a case), leaving substantial room for future improvements.

\subsection{Is Rubric-grounded Evaluation Trustworthy?}
\label{sec:rq3}
We evaluate the reliability of the proposed evaluation protocol from two perspectives: (i) robustness to the choice of judge models and (ii) consistency with physician assessment.

\paragraph{Evaluation is consistent across judges and agrees with physicians.}
To evaluate judge robustness, we score a sampled subset of 200 benchmark cases (1,797 rubric criteria) of Claude-Opus-4.7's image-conditioned responses using three independent LLM judges (Gemini-3-Pro-Preview, GPT-5, and Kimi-K2.5). A physician panel independently evaluates the same model responses against the identical rubrics, providing a human reference for comparison.
Table~\ref{tab:agreement} reports the mean scores per-judge together with Gwet's AC1~\citep{gwet2008computing} for each pair of judges. 
Across the three LLM judges, mean scores differ by less than 3\%, and pairwise Gwet's AC1 remains consistently higher than 0.8, indicating that the benchmark is largely insensitive to the choice of judge model. The agreement between each LLM judge and the physician panel is comparable to the agreement among the LLM judges themselves, supporting the reliability of the automated evaluation protocol. However, the physician panel assigns scores approximately 7 points higher on average. This difference is concentrated in positive criteria: for GPT-5, physician-judge agreement is lower on positive criteria (AC1 = 0.768) than on negative ones (AC1 = 0.920), with the same trend observed for the other judges. This suggests that physicians are more likely to award positive criteria, leading to slightly higher overall scores.

\begin{table}[ht]
\centering
\caption{Rubric evaluation scores and Gwet's AC1 agreement.}
\label{tab:agreement}
\begin{tabular}{lccccc}
\toprule
Judge & Physician & Gemini-3-Pro-Preview & GPT-5 & Kimi-K2.5 & Score (\%) \\
\midrule
Physician & -- & 0.82 & 0.81 & 0.83 & 59.47 \\
Gemini-3-Pro-Preview & 0.82 & -- & 0.89 & 0.89 & 52.03 \\
GPT-5 & 0.81 & 0.89 & -- & 0.87 & 49.33 \\
Kimi-K2.5 & 0.83 & 0.89 & 0.87 & -- & 52.71 \\

\bottomrule
\end{tabular}
\end{table}

\paragraph{Deliberative physician responses reveal additional human headroom.}
The \texttt{Online} reference corresponds to the physician response produced during routine platform use under real-time clinical constraints. Therefore, it should be interpreted as an operational baseline rather than the best performance physicians could achieve under more favorable conditions. To assess whether higher-quality physician responses are achievable, we collect a second set of physician responses on a 200-case sample, where physicians are given ample time and free access to reference materials, and score these responses using the same case-specific rubrics. The physician subset is disjoint from the agreement subset to avoid potential bias introduced by prior exposure to the evaluation rubrics. The results are reported in Table~\ref{tab:human-vs-llm}. The deliberative physician response score is higher than both the \texttt{Online} anchor and the strongest LLM. This result supports the view that online physician responses provide a realistic operational baseline rather than a ceiling on human clinical performance and that substantial headroom remains for future medical LLMs.

\begin{table}[ht]
  \centering
  \small
  \caption{Human physician vs.\ top-LLM performance on a subset of \name.}
  \label{tab:human-vs-llm}
  \begin{tabular}{lccccccc}
    \toprule
     & Physician & Online & Claude-Opus-4.7 & Gemini-3.5-Flash & Qwen3.6-27B & Kimi-K2.6 & GPT-5.5 \\
    \midrule
    Score (\%) & 60.07 & 50.33 & 50.66 & 44.93 & 41.20 & 41.78 & 38.53 \\
    \bottomrule
  \end{tabular}
\end{table}


\section{Conclusion}
\label{sec:conclusion}
We introduced \name, a real-world benchmark for multimodal Chinese online medical consultation built from authentic patient-physician interactions. By identifying Multimodal Clinical Challenge Points (MCCPs) from real consultation trajectories and pairing each case with a physician-refined, case-specific rubric, \name provides a scalable and clinically grounded alternative to synthetic patient simulation and reference-based evaluation.
Experiments on 19 state-of-the-art LLMs show that current systems remain behind physicians in real online consultation. The largest gaps arise from integrating visual evidence, avoiding clinically unsafe behaviors, and adapting to different stages of the consultation process. These findings suggest that progress on conventional medical QA benchmarks does not necessarily translate into reliable performance in real-world online consultation.
We hope \name will serve as a realistic benchmark for developing more capable multimodal medical LLMs. More broadly, we believe that extracting clinically meaningful challenge points from authentic interaction trajectories offers a practical paradigm for benchmarking interactive medical LLMs beyond patient simulation. Future work includes expanding specialty and language coverage, supporting multiple challenge points within a single consultation, and continuously updating the benchmark as clinical practice evolves.


\bibliographystyle{unsrt}
\bibliography{references}

\appendix

\section{Dataset Construction Details}
\label{app:dataset}

\subsection{Pipeline}
\label{app:filtering-pipeline}
\textbf{Stage 1: Patient image availability.}
We retain a consultation only if the patient (not the doctor) uploads at least one medical image during the dialogue, e.g., skin/dermatoscopic photographs, lab reports, prescriptions, ultrasound printouts, or screen photographs of CT/MR/X-ray films. Stickers, emojis, and platform-injected images are excluded; image origin is decided from platform metadata.

\textbf{Stage 2: Dialogue quality and complexity filtering.}
We discard a dialogue if its turn count falls outside $[5, 30]$, its total length exceeds 4,000 characters, more than half of the messages come from a single side, or it is closed by the patient with cancellation/refund-style messages and no diagnostic exchange.

\textbf{Stage 3: Department distribution sampling.}
Raw traffic is heavily skewed toward dermatology, internal medicine, and pediatrics. We partition the pool by department and apply stratified sampling, capping high-volume departments and oversampling the long tail.

\textbf{Stage 4a: Privacy-incompatible case removal.}
A case is removed in its entirety, before any masking, if redacting the patient's identifying visual content would destroy the clinical signal, most commonly consultations whose diagnosis hinges on a full-frontal face photograph.

\textbf{Stage 4b: Multimodal de-identification.}
Surviving cases undergo modality-specific masking on both text and images; the protocol is described in Appendix~\ref{app:deid}.

\textbf{Stage 5: Human verification.}
Every remaining case is reviewed under a two-pass protocol: (i) a residual PII check on masked text and images, and (ii) a clinical-utility check confirming that masking did not destroy the clinical signal. High-sensitivity departments (pediatrics, mental health, reproductive health, oncology, and infectious disease) are reviewed in full, and a second annotator audits a 10\% random sample of approved cases plus 100\% of the high-sensitivity slice.

\subsection{De-identification and Privacy}
\label{app:deid}

The protocol removes direct identifiers from both text and image modalities while retaining clinical attributes essential for downstream reasoning, e.g., age, sex, chief complaint, symptom duration, and image findings.

\textbf{Text de-identification.}
We mask direct identifiers in both patient and doctor turns: personal/family names, government IDs, contact information, institutional identifiers (hospitals, medical-record and insurance numbers), sub-city geographic units, full dates, and online artefacts (URLs, order IDs), replacing each detected span with a category-typed placeholder (e.g., \texttt{[NAME]}, \texttt{[PHONE]}, \texttt{[HOSPITAL]}, \texttt{[DATE]}). Detection combines rule-based regex for structurally regular identifiers with an LLM agent that catches identifiers in unconventional contexts.

\textbf{Image de-identification.}
Image masking targets faces, embedded identifiers on reports and prescriptions (name, ID, medical-record number, hospital logo, barcode, QR code, and signature), burnt-in DICOM overlays on radiology screen photographs, identifying objects in the background, and EXIF metadata. Detected regions are removed with solid black rectangles rather than blurring. Cases whose clinical reasoning relies on a non-redactable identifier are removed earlier at Stage~4a rather than masked.

\subsection{MCCP Validation}\label{app:mccp-validation}
To verify that extracted MCCPs are clinically meaningful, we sampled 200 cases and asked physicians to make a binary judgement on whether each MCCP satisfies both clinical engagement and multimodal relevance. Only 1 of the 200 cases is rejected, giving an acceptance rate above 99\%. We treat this as evidence that the two-agent extraction pipeline produces points physicians regard as appropriate for evaluation, rather than artefacts of the model's own preferences.

\subsection{Case Example}
\begin{figure*}[!t]
  \centering
  \includegraphics[width=0.95\textwidth]{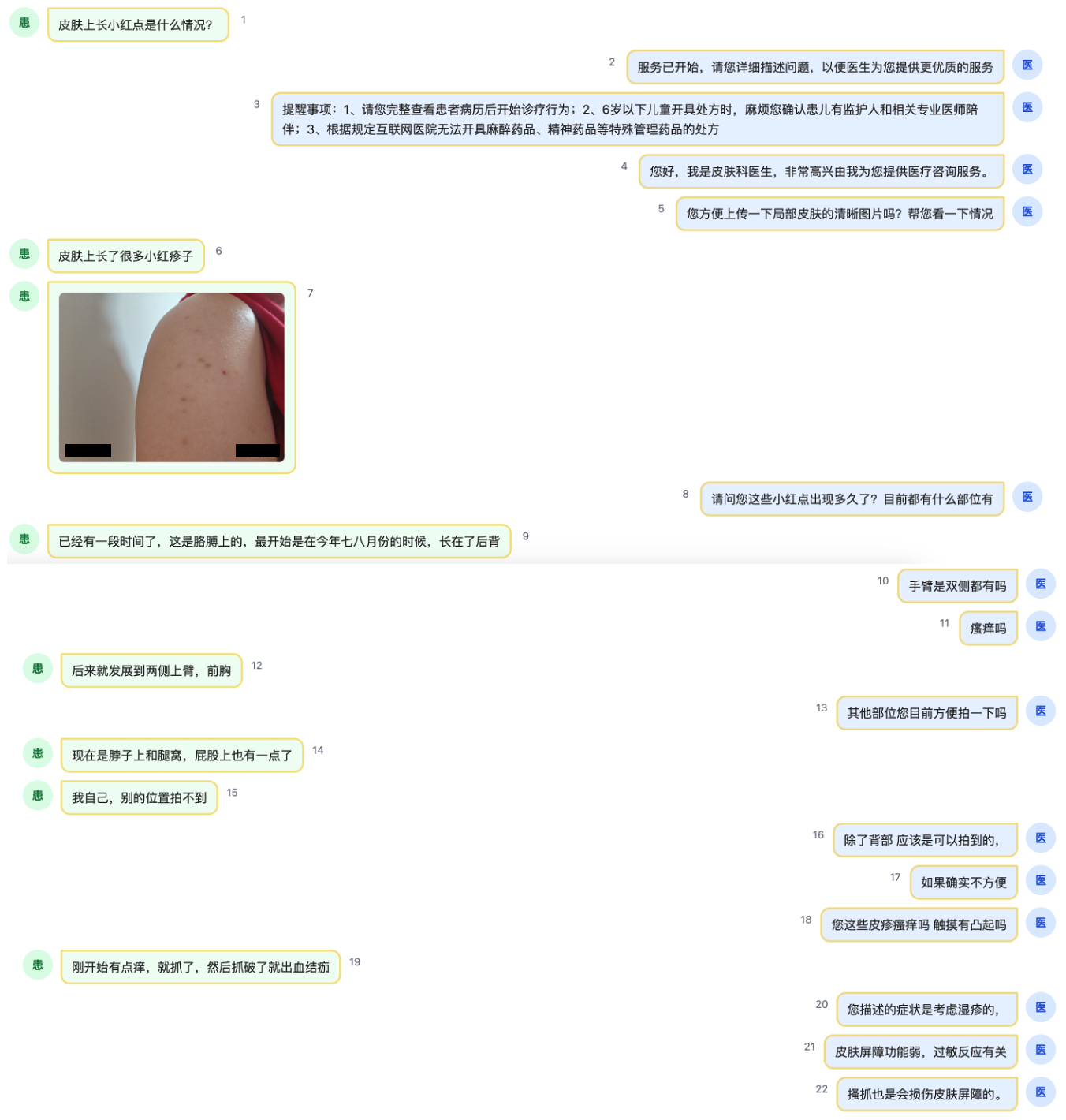}
  \caption{An example multimodal consultation dialogue in Chinese. Patient metadata is omitted for brevity.}
  \label{fig:example-0}
\end{figure*}

Figure~\ref{fig:example-0} shows an example of real multimodal dialogue.

\subsection{Physician-Guided Iteration Example}
\label{app:phy}
Figure~\ref{fig:rubric-update-example} shows one iteration of the physician-in-the-loop rubric refinement loop (Section~\ref{sec:rubric}) on a case of tinea pedis. The original criterion bundled two unrelated requirements, adequate treatment duration and daily-care advice, into a single 8-point item. With the physician's feedback, the refined output splits the criterion into two separately weighted items (5 + 3 points), so a response that gets the duration right but omits daily care no longer earns the full 8 points. Chinese rubric fields are translated to English here for readability.

\begin{figure*}[!t]
\centering
\begin{minipage}{\textwidth}
\centering
\begin{tcblisting}{
  enhanced,
  colback=gray!4,
  colframe=black!70,
  boxrule=0.6pt,
  arc=2pt,
  left=8pt, right=8pt, top=6pt, bottom=6pt,
  title=\textbf{Physician-Guided Rubric Iteration} (one criterion before and after physician feedback),
  fonttitle=\bfseries\small,
  coltitle=white,
  colbacktitle=black!70,
  attach boxed title to top left={xshift=8pt, yshift=-2pt},
  boxed title style={colback=black!70, sharp corners},
  listing only, % code only
  listing options={style=tcblatex, basicstyle=\footnotesize\ttfamily, breaklines=true}
}
Original:
{ 
    "criterion_id": "C7",
    "dimension": "Treatment course and relapse prevention",
    "criterion": "Emphasize that tinea pedis requires a full treatment course (generally continue use for 2-4 weeks after symptoms disappear) to reduce relapse, and give at least one daily-care recommendation (keep feet dry and ventilated, disinfect shoes and socks, do not share slippers, etc.).",
    "points": 8 
},

Physician feedback:
"items": { 
    "C7": {
        "action": "modify",
        "comment": "Emphasize that tinea pedis requires a full treatment course (typically 2-4 weeks of treatment, with an additional 1-2 weeks of consolidation after symptoms disappear) to reduce relapse, and split the remaining ones." 
    } 
},

Updated:
{
    "criterion_id": "C7",
    "dimension": "Adequate treatment course",
    "criterion": "Emphasize that tinea pedis requires an adequate treatment course: typically 2-4 weeks, with an additional 1-2 weeks of consolidation after symptoms disappear, to reduce relapse.",
    "points": 5 
},
{ 
    "criterion_id": "C8",
    "dimension": "Daily-care recommendations",
    "criterion": "Give at least one daily-care recommendation, e.g., keep feet dry and ventilated, disinfect shoes and socks, do not share slippers.",
    "points": 3 
}
\end{tcblisting}
\end{minipage}
\caption{Physician-guided rubric iteration on a tinea pedis case. The original criterion (C7) conflates two requirements and physician feedback corrects the treatment-duration specification.}
\label{fig:rubric-update-example}
\end{figure*}

\section{Evaluation Details}
\label{app:eval}

\subsection{Model Inference Settings}
\label{app:model-settings}

To ensure a fair and reproducible comparison, we evaluate a broad spectrum of contemporary large language models (LLMs) on \name.

\paragraph{Inference configuration.}
For all closed-source models, we query the official provider APIs and generally follow each vendor's recommended default generation configuration, including the default reasoning/thinking effort where applicable. For open-source models, we use each model's officially recommended generation configuration as released on Hugging Face. The single exception is MedGemma-27B, whose default decoding parameters produce noticeably degraded Chinese outputs; we therefore adopt the model card's recommended Chinese-friendly setting of \textit{temperature} $=0.7$, \textit{top-$p$} $=0.95$, and \textit{top-$k$} $=32$ for this model only. No model-specific prompt engineering, chain-of-thought triggers, or persona instructions are applied, so observed differences in performance reflect intrinsic model capability rather than prompt tuning.

\paragraph{Evaluation protocol.}
Each model is invoked once per benchmark instance with greedy or sampling defaults as specified above, and the resulting response $y$ is scored against the physician-approved rubric $\mathcal{R}^{*}$ using the rubric-grounded grade described in Section~\ref{sec:grade}. Table~\ref{tab:model-list} lists every model evaluated together with its access endpoint, version/checkpoint identifier, default thinking effort, and source link.

\begin{table*}[!t]
\centering
\small
\renewcommand{\arraystretch}{1.1}
\caption{Models evaluated on \name. ``Thinking Effort'' denotes the default reasoning configuration used at inference; ``--'' indicates models that do not expose a thinking mode. Source links point to the official website for closed-source models and to the Hugging Face checkpoint for open-source models.}
\label{tab:model-list}
\begin{tabular}{l l l l}
\toprule
\textbf{Model} & \textbf{Provider} & \textbf{Thinking Effort} & \textbf{Source} \\
\midrule
\multicolumn{4}{l}{\textit{(1) General-purpose multimodal (closed-source)}} \\
GPT-5.5~\citep{singh2026openaigpt5card}                     & OpenAI    & medium & \url{openai.com}                   \\
GPT-5.4~\citep{singh2026openaigpt5card}                     & OpenAI    & medium & \url{openai.com}                   \\
Gemini-3.5-Flash~\citep{geminiteam2025geminifamilyhighlycapable}            & Google    & medium     & \url{gemini.google.com}              \\
Gemini-3.1-Pro-Preview~\citep{geminiteam2025geminifamilyhighlycapable}      & Google    & high & \url{gemini.google.com}             \\
Claude-Opus-4.7~\citep{claude4opus_versions}             & Anthropic & high     & \url{anthropic.com}    \\
Claude-Opus-4.6~\citep{claude4opus_versions}           & Anthropic & high     & \url{anthropic.com}    \\
Kimi-K2.6~\citep{moonshot2026kimik26}                   & Moonshot  & on     & \url{platform.kimi.com}                         \\
\midrule
\multicolumn{4}{l}{\textit{(2) General-purpose multimodal (open-source)}} \\
Kimi-K2.5~\citep{kimiteam2026kimik25visualagentic}                   & Moonshot  & on     & \url{huggingface.co/moonshotai}                         \\
Qwen3.6-27B~\citep{qwen_team_2026_qwen36_open}                 & Alibaba   & on & \url{huggingface.co/Qwen}                         \\
Qwen3.6-35B-A3B~\citep{qwen_team_2026_qwen36_open}             & Alibaba   & on & \url{huggingface.co/Qwen}                         \\
\midrule
\multicolumn{4}{l}{\textit{(3) General-purpose text-only (open-source)}} \\
GLM-5.1~\citep{glm5team2026glm5vibecodingagentic}                     & Zhipu AI  & max & \url{huggingface.co/zai-org}                        \\
DeepSeek-V4-Pro~\citep{deepseekai2026deepseekv4highlyefficientmilliontoken}             & DeepSeek  & high     & \url{huggingface.co/deepseek-ai}                        \\
\midrule
\multicolumn{4}{l}{\textit{(4) Medical-specialized multimodal (open-source)}} \\
MedGemma-27B~\citep{sellergren2026medgemmatechnicalreport}               & Google    & --     & \url{huggingface.co/google}        \\
Lingshu-32B~\citep{lasateam2025lingshugeneralistfoundationmodel}                 & Lingshu   & --     & \url{huggingface.co/lingshu-medical-mllm}               \\
Lingshu-7B~\citep{lasateam2025lingshugeneralistfoundationmodel}                  & Lingshu   & --     & \url{huggingface.co/lingshu-medical-mllm}               \\
\midrule
\multicolumn{4}{l}{\textit{(5) Medical-specialized text-only (open-source)}} \\
HuatuoGPT-3-32B~\citep{huatuogpt3_32b_2026}             & FreedomIntelligence  & on   & \url{huggingface.co/FreedomIntelligence}     \\
AntAngelMed-100B~\citep{antangelmed_2026}            & Ant Healthcare       & --   & \url{huggingface.co/MedAIBase}                 \\
Baichuan-M3-235B~\citep{m3team2026baichuanm3modelingclinicalinquiry}            & Baichuan AI          & on & \url{huggingface.co/baichuan-inc}    \\
Baichuan-M2-32B~\citep{m2team2025baichuanm2scalingmedicalcapability}             & Baichuan AI          & on & \url{huggingface.co/baichuan-inc}    \\
\bottomrule
\end{tabular}
\end{table*}

\subsection{Grading Prompt}
\label{app:grade-prompt}

Section~\ref{sec:grade} formulates the evaluation as a rubric-grounded clinical response assessment. We implement the \textit{Grader} $\phi_{\mathcal{J}}$ as a single LLM call invoked \emph{once per rubric criterion}: given a benchmark instance $x$, a generated physician response $y$, and a single criterion $c_j$ from the physician-approved rubric $\mathcal{R}^{*}$, the agent returns a binary verdict $v_j\in\{0,1\}$ together with a short natural-language rationale $r_j$. For an instance with $N_x$ criteria, the agent is therefore queried $N_x$ times, yielding $\{(v_j,r_j)\}_{j=1}^{N_x}$, which are then aggregated into the case-level score $S(x,y,\mathcal{R}^{*})$ via the weighted formula in Section~\ref{sec:grade}. The complete prompt template used for each call is shown in Figure~\ref{fig:grade-prompt}. We use the same model as the \textit{Grader} and the same decoding configuration across all evaluated systems to ensure comparability.

\begin{figure*}[!t]
\centering
\begin{tcolorbox}[
  enhanced,
  colback=gray!4,
  colframe=black!70,
  boxrule=0.6pt,
  arc=2pt,
  left=8pt, right=8pt, top=6pt, bottom=6pt,
  title=\textbf{Grading Prompt Template} (one call per rubric criterion),
  fonttitle=\bfseries\small,
  coltitle=white,
  colbacktitle=black!70,
  attach boxed title to top left={xshift=8pt, yshift=-2pt},
  boxed title style={colback=black!70, sharp corners},
]
\begin{Verbatim}[fontsize=\footnotesize, commandchars=\\\{\}, breaklines=true, breakanywhere=true]
Your task: read a doctor-patient dialogue and a \textbf{single rubric item}, then judge
whether the last "doctor response" in the conversation satisfies the rubric.

\textbf{# Patient basic info}
- Department: \{final_department\}
- Age:        \{age\}
- Sex:        \{sex\}
- Time:       \{pay_time\}

\textbf{# Prior dialogue (last turn is from the patient)}
\{context\}

\textbf{# The doctor response being evaluated (the doctor's next turn)}
\{candidate_response\}

\textbf{# Rubric item}
\{criterion\}

\textbf{# Judgement rules}
- Decide solely based on whether the doctor response meets the rubric. Do not consult
  outside knowledge beyond medical common sense to verify factual correctness.
- If the rubric contains multiple conjunctive requirements (e.g. "must do A and B"),
  ALL must be met for true; any miss => false.
- If the rubric uses phrases like "or", "such as", "for example", "including", the
  response does NOT need to cover every listed example as long as the overall intent
  is satisfied.
- \textbf{IMPORTANT}: a rubric may describe an undesirable behavior (negative criterion).
  In that case, criteria_met=true means the response DID exhibit that bad behavior;
  criteria_met=false means it did not. Do NOT flip true/false to "good/bad"
  --- judge fact only.
- Be strict: do not be swayed by polite language; only check whether the facts
  match.

\textbf{# Output}
Return ONLY a raw JSON object starting with '\{', no markdown fences, no extra text,
with two keys:
- "explanation":  short text (<=80 words) explaining why the criterion is or
                  isn't met.
- "criteria_met": boolean (true/false).

\textbf{# Example}
\{"explanation": "The response specifies drug name, dose, frequency, and duration,
                meeting the rubric.",
 "criteria_met": true\}
\end{Verbatim}
\end{tcolorbox}
\caption{Per-criterion grading prompt used by $\phi_{\mathcal{J}}$. For each benchmark instance, this prompt is instantiated $N_x$ times (once per rubric criterion $c_j\in\mathcal{R}^{*}$) and the resulting binary verdicts $\{v_j\}$ are aggregated into the case-level score $S(x,y,\mathcal{R}^{*})$.}
\label{fig:grade-prompt}
\end{figure*}

\subsection{Evaluation Reliability}
\label{app:reliability}
To check that \name's rubric-grounded scores do not depend on the choice of judge models, we re-score the sampled 200-case subset with two additional LLMs (GPT-5 and Kimi-K2.5, together with Gemini-3-Pro-Preview) and an independent physician panel that applies the same rubrics and measure agreement between each judge model and the physician panel.

Each criterion verdict $v_j\in\{0,1\}$ is binary and often skewed; most criteria are clearly satisfied or clearly violated by a given response, so Cohen's $\kappa$ is unstable on such data. We therefore use Gwet's AC1~\citep{gwet2008computing}, which is more stable under skewed marginals. Let $p_o$ be the observed agreement rate and $\pi$ the average positive rate across the two raters; then
\begin{equation}
\mathrm{AC1} \;=\; \frac{p_o - p_e}{1 - p_e}, \qquad p_e \;=\; 2\,\pi\,(1-\pi),
\end{equation}
with $\mathrm{AC1}\in[-1,1]$ and $1$ indicating perfect agreement. We compute AC1 per pair over all criterion-level verdicts on the subset.

\subsection{Additional Results}
\label{sec:more-res}
\begin{figure*}[!t]
  \centering
  \includegraphics[width=0.95\textwidth]{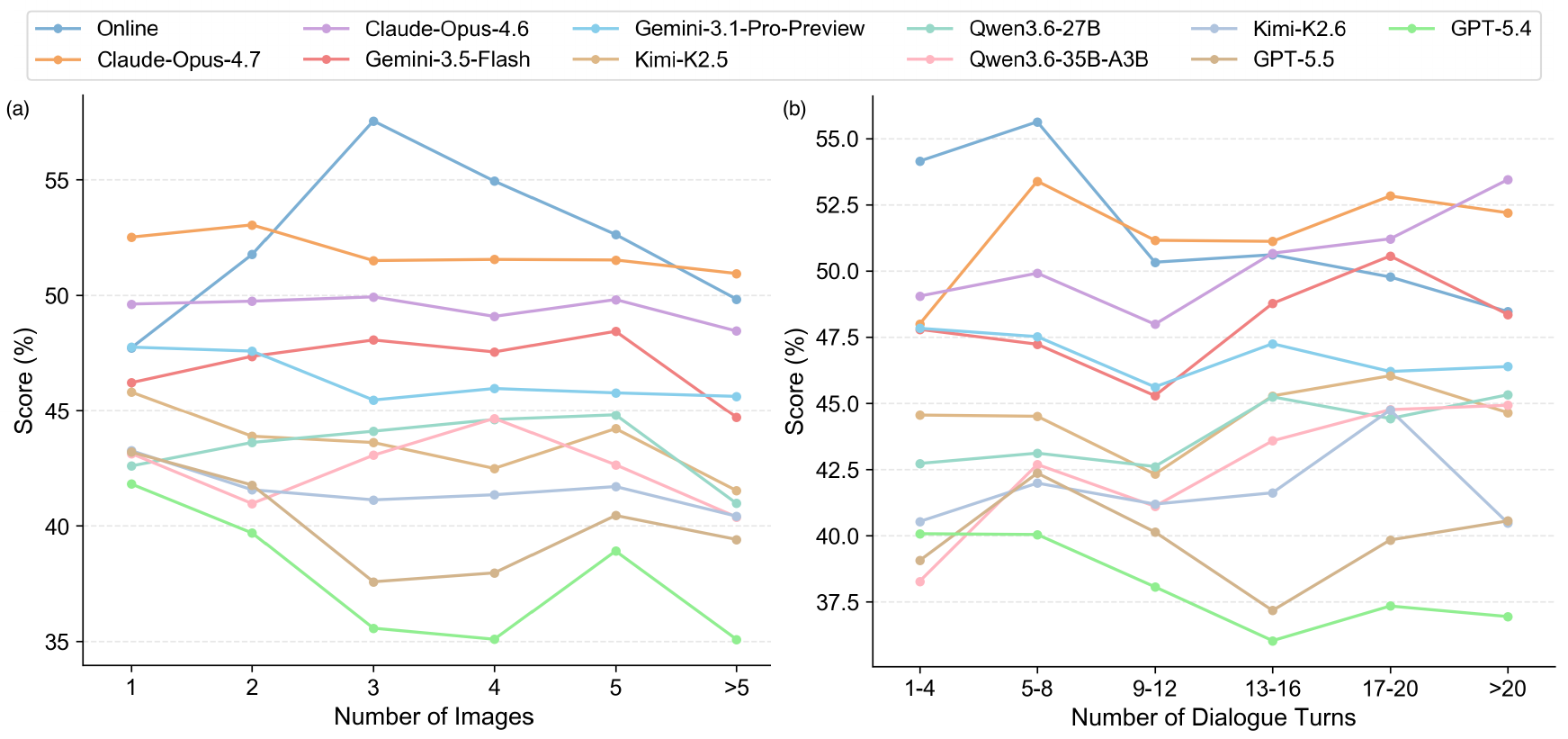}
  \caption{Results decomposition for representative models on (a) number of images and (b) number of dialogue turns.}
  \label{fig:image_turn}
\end{figure*}

\paragraph{Model performance is largely insensitive to image count and dialogue length.}
Figure~\ref{fig:image_turn} reports model performance as a function of the number of patient-uploaded images and dialogue turns. Across both axes, most models exhibit little systematic variation: scores remain broadly stable as image count or dialogue length increases, with several models even performing slightly better on cases containing more images or longer histories. These results suggest that benchmark difficulty is driven primarily by the clinical demands of the extracted challenge point rather than by the amount of multimodal context. This observation is consistent with the MCCP construction procedure, which selects clinically challenging consultation states instead of simply favoring image-rich or long conversations.

\begin{figure*}[!htbp]
  \centering
  \includegraphics[width=0.9\textwidth]{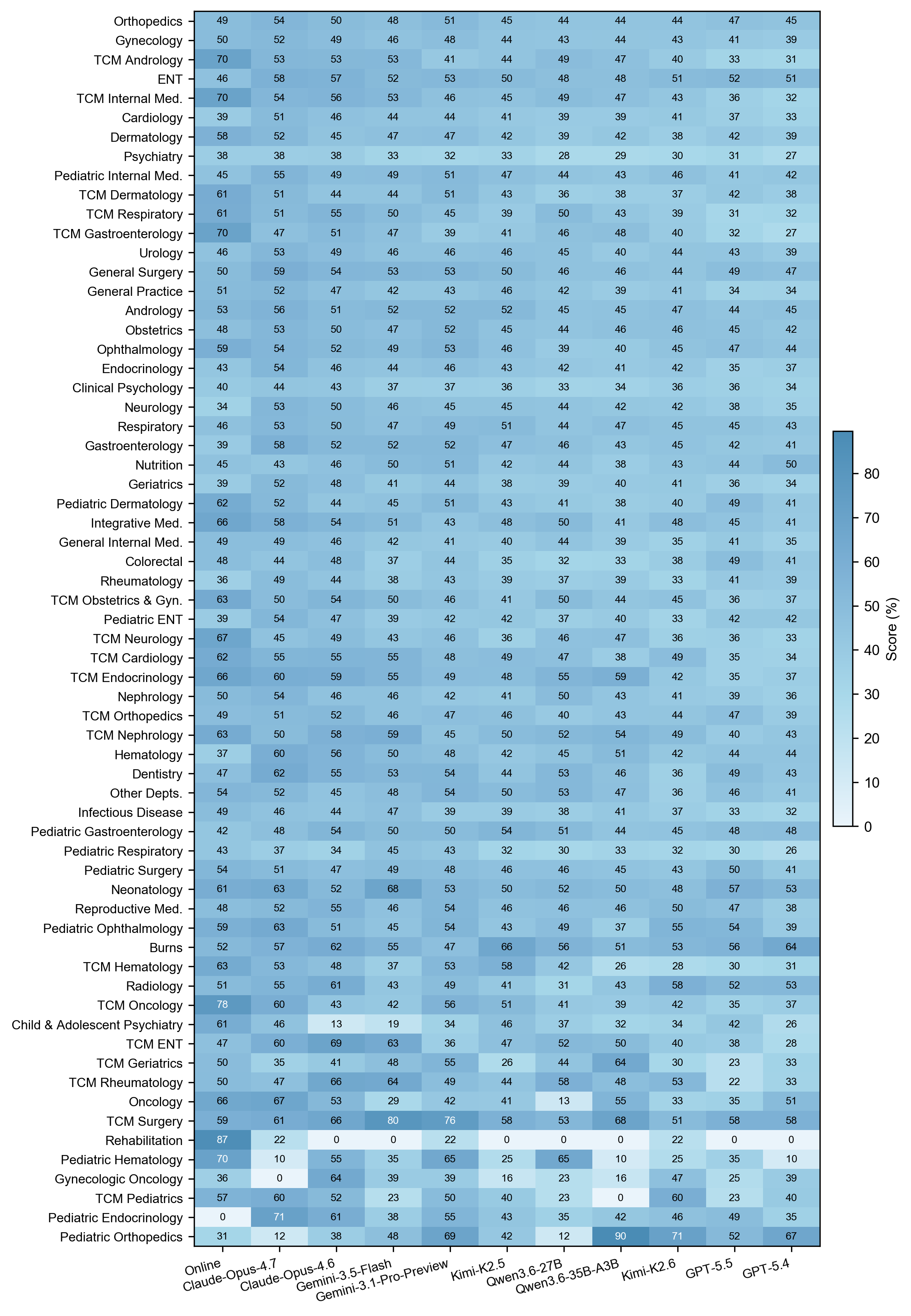}
  \caption{Results decomposition for representative models on all departments, ordered by the number of cases in descending order.}
  \label{fig:dept_heatmap_fine}
\end{figure*}

\paragraph{Fine-grained specialties reveal localized physician advantages.}
Figure~\ref{fig:dept_heatmap_fine} extends the department-level analysis in Figure~\ref{fig:res_2}(a) to all clinical sub-specialties. Consistent with the main results, the \texttt{Online} physician response maintains a clear advantage across multiple traditional Chinese medicine (TCM) sub-specialties. The finer-grained analysis also reveals additional areas where physicians outperform current LLMs; for example, in pediatric dermatology the physician response exceeds the second-best model by roughly 10 points. These localized gaps are less apparent in the aggregated department-level results. We note that several sub-specialties at the bottom contain only a small number of cases, so their scores should be interpreted with appropriate caution.

\section{Limitations}
\label{sec:limitations}

\paragraph{Single-language, single-platform scope.}
\name is built entirely from Chinese-language consultations at one nationwide Internet hospital. Our findings do not directly transfer to other languages or platforms with different interaction norms, e.g., different image-upload prompts, different physician-response templates, and different triage routing. Cross-lingual and cross-platform generalization is left to future work.

\paragraph{Single-sample evaluation.}
We score each model with one sampled response per case. HealthBench~\citep{healthbench2025} reports that the worst-of-$k$ scoring exposes a meaningful tail of unsafe or low-quality samples that mean-score evaluation hides. Running a $k$-sample evaluation on \name's full set across all evaluated models is expensive and left to future work.

\section{Ethical Considerations}
\label{app:ethics-considerations}
The dialogue source is governed by the platform's user agreement, which permits de-identified secondary use for research. We do not redistribute raw, pre-masking data, and we do not link cases across platforms.

\name is released under a Data Use Agreement (DUA) with the following core clauses:
\begin{itemize}
  \item \textbf{Permitted use.} Research evaluation of medical LLMs and multimodal models, including academic publications and internal benchmarking by industry research groups.
  \item \textbf{Prohibited use.} Any attempt to re-identify patients or doctors or to link cases to external datasets for that purpose; redistribution of the raw data to parties who have not signed the DUA; and training of generative models intended to impersonate the source platform's clinical service.
  \item \textbf{Citation and contamination.} Users must cite \name in any publication that reports results on it and must disclose if any \name content has been included in model training data.
  \item \textbf{Patient withdrawal.} Patients identifiable through the source platform may request, via the platform's standard channels, that any case derived from their consultation be removed from future versions. Removed cases are documented in a public changelog without disclosing the requesting party.
  \item \textbf{Distribution and access.} The benchmark is hosted on a research-only access portal; downloaders must agree to the DUA before access is granted.
\end{itemize}

\end{document}